\documentclass{article}
\usepackage{ijcai16}

\usepackage{times}
\usepackage{helvet}
\usepackage{courier}
\usepackage{url}
\usepackage{latexsym}
\usepackage{multirow}
\usepackage{float}
\usepackage{stfloats}
\usepackage{amssymb,amsmath,mathrsfs,graphicx,epsfig,epstopdf}

\usepackage{bm}
\usepackage{color}
\usepackage{array}
\usepackage{caption}
\usepackage{lineno}
\usepackage{slashbox}

\usepackage{CJKutf8}

\begin{document}
\title{Chinese Song Iambics Generation with Neural Attention-based Model}
\author{
Qixin Wang$^{1,4}$\footnotemark[1], Tianyi Luo$^{1,3}$\footnotemark[1], Dong Wang$^{1,2}$\footnotemark[2], Chao Xing$^{1}$\\
  $^1$CSLT, RIIT, Tsinghua University, China \\
  $^2$Tsinghua National Lab for Information Science and Technology, Beijing, China\\
  $^3$Huilan Limited, Beijing, China \\
  $^4$CIST, Beijing University of Posts and Telecommunications, China \\
  {\tt \{wqx, lty, xingchao\}@cslt.riit.tsinghua.edu.cn} \\
  {\tt wangdong99@mails.tsinghua.edu.cn}
  \\
  }
\maketitle
\footnotetext[1]{The two authors contributed equally.}
\footnotetext[2]{Corresponding author: Dong Wang; RM 1-303, FIT BLDG, Tsinghua University, Beijing (100084), P.R. China.}
\begin{abstract}
  Learning and generating Chinese poems is a charming yet challenging task. Traditional approaches involve various language modeling and machine translation techniques, however, they perform not as well when generating poems with complex pattern constraints, for example Song iambics, a famous type of poems that involve variable-length sentences and strict rhythmic patterns.

  This paper applies the attention-based sequence-to-sequence model to generate Chinese Song iambics. Specifically, we encode the cue sentences by a bi-directional Long-Short Term Memory (LSTM) model and then predict the entire iambic with the information provided by the encoder, in the form of an attention-based LSTM that can regularize the generation process by the fine structure of the input cues. Several techniques are investigated to improve the model, including global context integration, hybrid style training, character vector initialization and adaptation. Both the automatic and subjective evaluation results show that our model indeed can learn the complex structural and rhythmic patterns of Song iambics, and the generation is rather successful.
\end{abstract}

\section{Introduction}

The classical Chinese poetry is an important and special cultural heritage with over 2,000 years of history. There are many genres of Chinese classical poetry, including Tang poetry, Song iambics, Ming poetry and Qing poetry. Different types of classical Chinese poetry possess their own specific structural, rhythmical and tonal patterns. The structural pattern regulates how many lines and how many characters for each line; the rhythmical pattern requires that the last characters of certain lines hold similar vowels; the tonal pattern requires characters in particular positions hold particular tones, i.e., `Ping'(level tone), or `Ze'(downward tone). A good poem should follow all these three pattern regulations (in a descendant order of priority), and has to hold consistent semantic meaning and emotional characteristics. For this reason, it is very difficult to generate Chinese classical poems even for people.

Roughly speaking, Chinese classical poetry can be classified into regulated verses and iambics. Regulated verses were mostly popular in Tang dynasty (therefore often called `Tang poetry'), and iambics gained the most popularity in Song dynasty (so often called `Song iambics', or `\begin{CJK*}{UTF8}{gbsn} 宋词\end{CJK*}' in Chinese). Compared to regulated verses that hold very strict structures (fixed number of lines and fixed number of characters per line) and rhythmical patterns, Song iambics are more flexible: their structures and rhythmical patterns are not necessarily identical, instead each lyric may follow one of some pre-defined `tunes'. Actually, Song iambics were originally lyrics of Songs performed by young female artists, which fostered different tunes to match different melodies. An example of Song iambics whose tune is `\begin{CJK*}{UTF8}{gbsn}虞美人\end{CJK*}(Beauty Yu)' is shown in Table~\ref{tab:Songsample}, where the rhythmical patterns are labelled as bold characters, and the tonal patterns are provided after each line, where `P' represents level tone and `Z' represents downward tone.

\begin{table}[!htb]
\begin{center}
\begin{tabular}{|c|c|c|}
\hline
\begin{CJK*}{UTF8}{gbsn}
虞美人
\end{CJK*}
\\
Beauty Yu\\
\begin{CJK*}{UTF8}{gbsn}
春花秋月何时\textbf{了}，\,\,\,\,(*\,P\,*\,Z\,P\,P\,Z)
\end{CJK*}
\\
Flowers bloom and wither, the moon rises and sets. \\
When can it end?\\
\begin{CJK*}{UTF8}{gbsn}
往事知多\textbf{少}。\,\,\,\,\,\,\,(*\,Z\,P\,P\,Z)
\end{CJK*}
\\
As for stories buried in the past, who will really attend?\\
\begin{CJK*}{UTF8}{gbsn}
小楼昨夜又东\textbf{风}，\,\,\,\,(*\,P\,*\,Z\,Z\,P\,P)
\end{CJK*}
\\
Wind blew over my attic last night,\\
\begin{CJK*}{UTF8}{gbsn}
故国不堪回首月明\textbf{中}。(*\,*\,*\,P\,*\,Z\,Z\,P\,P)
\end{CJK*}
\\
How is my home country now, in the same moonlight?\\
\begin{CJK*}{UTF8}{gbsn}
雕阑玉砌应犹\textbf{在}，\,\,\,\,(*\,P\,*\,Z\,P\,P\,Z)
\end{CJK*}
\\
I bet the jade banisters and steps \\
are as exquisite as they were,\\
\begin{CJK*}{UTF8}{gbsn}
只是朱颜\textbf{改}。\,\,\,\,\,\,\,(*\,Z\,P\,P\,Z)
\end{CJK*}
\\
I guess it is only the people who changed for sure.\\
\begin{CJK*}{UTF8}{gbsn}
问君能有几多\textbf{愁}，\,\,\,\,(*\,P\,*\,Z\,Z\,P\,P)
\end{CJK*}
\\
My sorrow,\\
\begin{CJK*}{UTF8}{gbsn}
恰是一江春水向东\textbf{流}。(*\,*\,*\,P\,*\,Z\,Z\,P\,P)
\end{CJK*}
\\
Flows like the river. It never ends.\\
\hline
\end{tabular}
\end{center}
\caption{An example of Song iambics with a popular tune `Beauty Yu'. The rhyming characters are in boldface, and the tonal pattern is shown at the end of each line, where `P' indicates level tone and `Z' indicates downward tone, and `*' indicates that the tone of this character can be either level or downward.}
\label{tab:Songsample}
\end{table}

In this paper, we are concerned with automatic generation for Song iambics, not only because of its practical value in entertainment and education, but also because it demonstrates an important aspect of artificial intelligence: the creativity of machines in art generation. Although some researches have been conducted for Chinese classical poetry generation, most of the existing approaches focus on Tang poetry, particularly quatrains. For more flexible genres such as Song iambics, little progress has been achieved. There are many difficulties in Song iambics generation compared to generating Tang poetry. Firstly, Song iambics are often much longer than Tang poetry, which makes it not easy to control the theme (e.g., topics, emotional status) and the semantic flow (i.e., relations between consecutive lines); secondly, Song iambics follow more complicated and diverse regulations in structures, rhythms and tones, which is not trivial to learn; thirdly, for most tunes, there is only a very limited number of Song iambics that have been passed down to this day, leading to difficulty for model training.

We propose a novel attention-based Long-Short Term Memory (LSTM) model for Song iambics generation. Specifically, we follow the sequence-to-sequence learning architecture, and use the LSTM model as the encoder and decoder. It is well-known that the LSTM model is capable of learning long-distance information and so can largely alleviate the quick-forgetting problem associated with the traditional RNN model~\cite{hochreiter1997long}. Additionally, the attention-based approach proposed recently by~\cite{bahdanau2014neural} was adopted to provide fine-grained supervision for the generation process. The attention-based approach generates each character by referring to all the characters of the input cue sentence, and automatically locates the most relevant character that the generation should be based on. This is a powerful supervision mechanism that enables accurate character-level supervision and thus can model the strict structural regulations and the subtle emotional states of Song iambics. Since the generation always looks back on the input sentence, the entire generation is strongly enforced to follow the same theme.  This is particular important for Song iambics generation that often suffers from severe `concept drift' when generating more than a few sentences.


\section{Related Work}

Poetry automatic generation is a challenging research topic over the past decades. The first approach is based on rules and templates. For example, \cite{tosa2009hitch,wu2009new} employed a phrase search approach for Japanese poetry generation, and~\cite{netzer2009gaiku} proposed an approach based on word association norms. \cite{oliveira2009automatic,oliveira2012poetryme} used semantic and grammar templates for Spanish poetry generation.

The second approach is based on various genetic algorithms~\cite{manurung2004evolutionary,manurung2012using,zhou2010genetic}. For example, \cite{zhou2010genetic} proposed to use a stochastic search approach to obtain the best matched sentences.

The third approach to poetry generation is by various statistical machine translation (SMT) methods. This approach was used by~\cite{jiang2008generating} to generate Chinese couplets, a type of simple regulated verses with only two lines.  \cite{he2012generating} extended this approach to generate Chinese quatrains (four-line Tang poems), where each line of the poem is generated by translation from the previous line.

Another approach to poem generation is based on summarization. For example, ~\cite{yan2013poet} proposed a method that retrieves high-ranking candidates of sentences out of a large poem corpus given users' queries.
These candidates are segmented into constituent terms which are then grouped into clusters.
By re-organizing the terms from different clusters iteratively, sentences that conform the regulation patterns are selected as the generation results.

More recently, deep learning methods gain much interest in poetry generation. For example, \cite{zhang2014chinese} proposed an RNN-based approach to generate Tang poems. By this approach, the first line is generated by a character-based RNN language model~\cite{mikolov2010recurrent} given some input keywords, and then the subsequent lines are generated sequentially by accumulating the status of the sentences that have been generated so far.

Our approach follows the RNN-based approach and thus closely related to the work~\cite{zhang2014chinese}. However, several important differences make our proposal novel. Firstly, we use the LSTM rather than the conventional RNN to obtain long-distance memory; secondly, we use the attention-based framework to enforce theme consistency; thirdly, our model is a simple sequence-to-sequence structure, which is much simpler than the model proposed by~\cite{zhang2014chinese} and can be easily extended to generate poems with various genres. Particularly, we employ this model to generate Song iambics that involve much more complex structures than Tang poems and have never been successfully generated by machines.

\section{Method}

In this section, we first present the attention-based Song iambics generation framework, and then describe the implementation of the encoder and decoder models that have been tailored for our task.

\subsection{Attention-based Song Iambics Generation}

The attention-based sequence-to-sequence learning provided by~\cite{bahdanau2014neural} is a general framework where the input sequence is converted to a sequence of hidden states that represent the semantic status at each position of the input, and these hidden states are then used to regulate the generation of the target sequence. The model that generates the hidden states is called `encoder', and the model that generates the target sequence is called `decoder'. The important mechanism of the attention-based model is that at each generation step, the most relevant input is discovered by comparing the `current' status of the decoder with the hidden states of encoder, so that the generation is regulated by the fine structure of the input sequence.

\begin{figure}[!htb]
\centering
\epsfig{figure=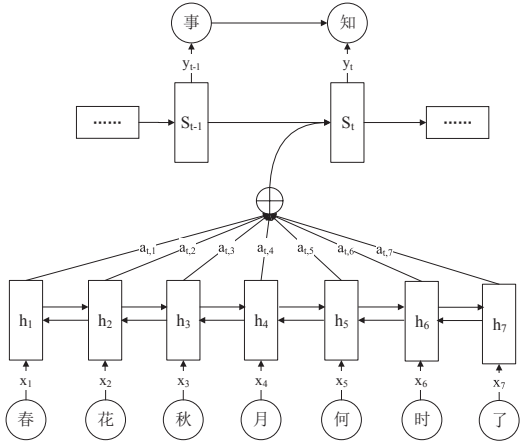,width=8cm}
\caption{The attention-based sequence-to-sequence learning framework for Song iambics generation.}
\label{fig:model}
\end{figure}

The entire framework of the attention-based model applied to Song iambics generation is illustrated in Figure~\ref{fig:model}. As for each iambic, the encoder (a bi-directional LSTM that will be discussed shortly) converts the input sentence (the first line of the iambic), a character sequence denoted by $(x_1, x_2, ...)$ where $x_i$ denotes the embedding representation of the $i$-th character, into a sequence of hidden states $(h_1,h_2,...)$.  The decoder then uses these hidden states to generate the remaining sentences in the iambic character by character, denoted by $(y_1,y_2,...)$. 
At each step $t$ of the generation, the prediction for the character $y_{t}$ is based on the `current' status $s_t$ of the decoder as well as the hidden states $(h_1,h_2,...)$ of the encoder. Importantly, each hidden state $h_i$ contributes to the generation controlled by a relevance factor $\alpha_{t,i}$ that measures the similarity between $s_t$ and $h_i$. By this mechanism, the decoder will pay more attention to the part of the input cue sentence that is mostly relevant to the current generation.

\subsection{LSTM-based Attention Model Structure}

A potential problem of the conventional RNN is that it tends to forget the historical input quickly, and so are not suitable to learn long-distance patterns that are often observed in Song iambics. To improve the capacity in memorizing long-distance patterns, we employ a bi-directional LSTM model as the encoder, which consists of two LSTMs that encode the input sequence in both forward and backward directions. It is well-known that LSTM is capable of learning long history, and using the bi-directional structure further improves this capability.

For the decoder, we use another LSTM. It maintains an internal status vector $s_t$, and for each generation step $t$, a context input $c_t$ is accepted and the most probable output $y_t$ is generated based on $s_t$. This can be formulated as follows:

\begin{eqnarray}
\nonumber
y_t = argmax_{y} p(y|y_{t-1},s_t,c_t).
\end{eqnarray}

\noindent After each prediction, $s_t$ is updated by

\[
s_t = f(s_{t-1},y_{t-1},c_{t})
\]

\noindent where $f(\cdot)$ is the update function that is determined by the model structure.

The context vector ${c}_{t}$ represents the external input during the generation, and is often used to provide some global information. For example in~\cite{zhang2014chinese}, $c_t$ is derived from all the sentences that have been generated so far. In the attention-based approach, $c_t$ is derived from all the hidden states of the input sequence, e.g., the first sentence provided by users, formulated as:

\begin{eqnarray}
\nonumber
{c}_{t} = \sum_{j=1}{\alpha }_{tj}{h}_{j},
\end{eqnarray}
\noindent where $h_j$ is the hidden state after the $j$-th input character is encoded, and $\alpha_{i,j}$ is the `attention' on $h_j$, derived by:

\begin{eqnarray}
\nonumber
{\alpha }_{ij}=\frac{exp({e}_{ij})}{\sum_{k=1}^{{T}_{x}}exp({e}_{ik})}
\end{eqnarray}
\noindent and
\begin{eqnarray}
\nonumber
a(s_{i-1},h_{j})=v_{a}^\mathrm{T}tanh(W_{a}+U_{a}h_{j}),
\end{eqnarray}
\noindent where $v_{a}$, $W$ and $U$ are three matrices that need to be optimized during model training.

\subsection{Model Training}

The goal of the model training is to let the predicted character sequence match the original Song iambics. We choose the cross entropy between the distributions over Chinese characters given by the decoder and the ground truth (essentially in a one-hot form) as the objective function. To speed up the training, the minibatch stochastic gradient descent algorithm is adopted. The gradient is computed sentence by sentence, and the AdaDelta algorithm is used to adjust the learning rate~\cite{zeiler2012adadelta}.

\section{Implementation}

The basic attention model, even with LSTMs, does not naturally work well for Song iambics generation. A particular problem is that there are more than 1000 tunes for Song iambics, and most of the tunes can find very limited number of works. On the other hand, the attention model is fairly complex and involves a large amount of free parameters. This makes the model training fairly difficult. We present several techniques to improve the model in this section.

\subsection{Global Context Supervision}

The global representation of the input sequence deserves special treatment, because only at the end of the sentence, the true intention of the sentence becomes clear. We concatenate the hidden states of the forward and backward LSTMs at their own last time steps as the sequence representation, and use it as the initial input of the generation. This strategy has been used in~\cite{cho2014learning,sutskever2014sequence} for building the intermediate representation for machine translation, and the method proposed by~\cite{zhang2014chinese} belongs to this category as well.


\subsection{Character Vector Initialization and Adaptation}

Due to the limited training data, we propose an initialization approach with character vectors.
We first derive character vectors using the word2vec tool\footnote{\url{https://code.google.com/archive/p/word2vec/}} based on a large external corpus, and then use these character vectors to initialize the word embedding matrix in the attention model. Since part of the model parameters (embedding matrix) have been pre-trained, the problem of data sparsity associated with the attention model can be alleviated.


We study two model training strategies with the character initialization. The first strategy (fixV) fixes the word vectors during the attention model training, while the second strategy (adaptV) adapts the word vectors together with other parameters. The second approach optimizes the attention model more aggressively, but may run the risk of over-fitting.

\subsection{Hybrid-tune Training}

For Song iambics, each tune holds its own regulation. This means that the dynamic property of each tune is unique and therefore should be modeled by different models, i.e., LSTMs in our attention model. However, the training data for most of the tunes is very limited, which means that training an individual model for each tune is almost impossible, except for a very few popular tunes.

We propose a hybrid-tune training approach to solve this problem. Basically, all the tunes share the same attention model, and a `tune indicator' is augmented to the context vector $c_1$ to notify the model which tune the training or the generation is processing. Specifically, it is added to the first hidden state of the LSTM decoder via a linear transform. In our study, the tune indicators are derived as eigen vectors of a $200 \times 200$ dimensional random matrix, and they are fixed during the model training and inference.

\section{Experimental Design}

\begin{table*}[tp]
\begin{center}
\begin{tabular}{|c|c|c|c|c|c|}
\hline
Algorithm              & \multicolumn{2}{|c|}{Fluency}               & \multicolumn{2}{|c|}{Meaingfulness}  &\,\,\,\,\,\,Total\,\,\,\,\,\,\\
 \cline{2-5}       & Partridge Sky     & Pusaman      &   Partridge Sky    & Pusaman & \\
\hline
{\bf{Song-fixV}}:Song-adaptV        & 63:{\bf69}     & {\bf74}:71      &   {\bf74}:58    & {\bf74}:71 &{\bf285}:269 \\
\hline
{\bf{Song-fixV}}:Giga-fixV          & {\bf83}:64      & {\bf75}:67  & {\bf75}:72 & 66:{\bf76} &{\bf299}:279 \\
\hline
Giga-fixV:{\bf{Song-adaptV}}     & 64:{\bf73}        &    49:{\bf79}    &   {\bf69}:68   & 46:{\bf82} & 228:{\bf302}\\
\hline
{\bf{Song-fixV+Global}}:Song-fixV     &  {\bf88}:59          & {\bf73}:69    &   {\bf77}:70   &67:{\bf75} &{\bf305}:273 \\
\hline
{\bf{Song-fixV+Global+Hybrid11}}:Song-fixV+GO     &  {\bf129}:112          & {\bf73}:69    &   {\bf123}:116   &{\bf120}:119 &{\bf445}:416 \\
\hline
\end{tabular}
\end{center}
\caption{\label{tab:bestresult} Comparison of the attention model with different configurations. `Song' and `Giga' denote the two databases used to train word vectors: Songci and Gigaword; `fixV' and 'adaptV' represent the fixed word vector strategy and the adapted word vector strategy respectively. `Global' represent the global context supervision approach, and `Hybrid11' denotes hybrid-tune training with Song iambics from $11$ tunes. }
\end{table*}

\begin{table*}[tp]
\begin{center}
\begin{tabular}{|c|c|c|c|c|c|c|c|}
\hline
Model       & \multicolumn{2}{|c|}{Poeticness}       & \multicolumn{2}{|c|}{Fluency }               & \multicolumn{2}{|c|}{Meaningfulness}  &\,\,\,\,\,Average\,\,\,\,\, \\
 \cline{2-7}    & Partridge Sky     & Pusaman   & Partridge Sky     & Pusaman      &   Partridge Sky    & Pusaman & \\
\hline
SMT    & 2.96     & 3.96    & 2.05     & 2.85      &   2.09    & 3.00 &2.82 \\
\hline
RNNLM       & 3.19     & 3.88   & 2.50      & 3.02  & 2.42 & 3.10 &3.02 \\
\hline
Attention-1   & 4.68     & {\bf4.98}  & 3.00        &    3.40    &   2.73   & 3.37 & {\bf3.69} \\
\hline
Attention-11   & {\bf4.71}     & 4.80  &  3.27          & 3.06    &   3.15   &3.22 &{\bf3.70} \\
\hline
Attention-All  & 4.68     & 4.92   &  3.52          & 3.88    &   3.60   &3.90 &{\bf4.08} \\
\hline
Human  & 4.31     & 4.77   &  {\bf4.67}          & {\bf4.23}    &   {\bf4.40}   &{\bf4.30} &{\bf4.45} \\
\hline
\end{tabular}
\end{center}
\caption{\label{tab:bestresult1} Averaged ratings for Song iambics generation with different methods and settings. `Partridge Sky' and `Pusaman' are the two tunes in the evaluation. }
\end{table*}

\begin{table}[!htb]
\begin{center}
\begin{tabular}{|l|c|c|}
\hline
Model                                & BLEU \\
\hline
SMT                                  & 0.0598\\
RNNLM                         & 0.0330\\
Attention-1                   & 0.0645\\
Attention-11                        & 0.0742\\
Attention-All                        & {\bf 0.1482}\\
\hline
\end{tabular}
\end{center}
\caption{\label{tab:bleu} BLEU-2 scores with different generation methods. }
\end{table}

This section presents the experimental study, particularly we compare the proposed model and two popular methods in poetry generation: the SMT-based approach and the LM-based approach. Note that we didn't see much work in Song iambics generation, and so had to implement the comparative approaches by ourselves, with the effort as much as we can pay.

\subsection{Data and Experiment Setup}

Several datasets are used to conduct the experiments. Firstly a Song iambics corpus (Songci) was collected from the Internet. This corpus consists of $15,689$ Song iambics in total. As far as we know, this covers most of the Song iambics that come down through the years. Among these iambics, $15,001$ are used for training and $688$ are used for test. The second dataset involves two corpora used to train the word embedding model, including the Gigaword corpus (contains roughly 1.12 billion Chinese characters) and the Songci corpus (1.22 million Chinese characters roughly).

The SMT model is phrase-based and was built using the Moses tool~\cite{koehn2007moses}. We found that using the Songci corpus only (using Moses) can not lead to reasonable performance. To improve the SMT approach, we use $11,099$ quatrains and $62,566$ other regulated verses, plus the Songci corpus to train the model.

For the attention model, both the encoder and decoder involve a recurrent hidden layer that contains $500$ hidden units, and a non-recurrent hidden layer that contains $600$ units. A max-out non-linear layer is then employed to reduce the
dimensionality to $300$, followed by a linear transform to generate the output units that correspond to the possible Chinese characters. The model is trained with the AdaDelta algorithm~\cite{zeiler2012adadelta}, where the minibatch is set to be $60$ sentences.

In all the comparative methods, the rhythmical and tonal patterns are strictly enforced by selecting a rule-compliant
character from the n-best candidates at each prediction step. If there are no candidates satisfying the rule, the top-1
candidate is selected.

\subsection{Human Evaluation}

The main evaluation is subjective and is conducted by experts. The evaluation was conducted in two phases. In the first phase, we focused on configuration selection, e.g., which corpus to use to train the word vectors and whether the word vectors should be adapted during the attention model training; in the second phase, we compared the attention model with the best configuration and the alternative approaches. We select two most popular tunes of Song iambics in the experiment, `Partridge Sky' and `Pusaman', for which there are $441$ and $403$ iambics respectively in the training data.

The evaluation is based on three metrics: poeticness (if the generated iambics follow the regulation on tone and rhyme), fluency (if the sentences are fluent and convey reasonable messages), and meaningfulness (if the entire generation focuses on a single theme and exhibits some consistent semantic meaning). We select $34$ `Partridge Sky' and $31$ `Pusaman' in the evaluation.

\subsection{BLEU Evaluation}

Bilingual Evaluation Understudy (BLEU)~\cite{papineni2002bleu} was originally proposed to evaluate machine translation systems. We follow~\cite{zhang2014chinese} and use the BLEU-2 score as the second evaluation metric for Song iambics generation, considering that most words in traditional Chinese consist of one or two characters. The method proposed by~\cite{he2012generating} and employed by~\cite{zhang2014chinese} was adopted to obtain reference iambics automatically. A slight difference is that the reference sets were constructed for each \emph{input cue sentence} in the test, instead for all the sentences in the test iambics. This is because our attention model generates iambics as an entire character sequence, instead of sentence by sentence in~\cite{zhang2014chinese}.

\section{Experimental Results}

We report the evaluation in two phases, where the first phase focuses on searching for optimal configurations for the attention model, and the second phase compares the attention model with other methods.

\subsection{Results in Phase 1}

In this evaluation, we intend to find the best configurations for the proposed attention-based model.
We invited 16 experts\footnote{These experts are professors and their postgraduate students in the field of Chinese Song Iambics research.  Most of them are from the Chinese Academy of Social Sciences (CASS).} to conduct a series of pair-wised evaluations, where the participants were asked to choose a better one from a pair of iambics produced by the attention model with different configurations. The votes for each configuration are aggregated to measure the goodness. Table~\ref{tab:bestresult} presents the results, where the numbers represent the number of votes in each pair-wised comparison.

From these results, one can observe that word vectors trained with the Songci corpus outperforms the word vectors trained with the Gigaword corpus. This is understandable since Sognci consists all the 15k Song iambics and so matches the domain of our task. Additionally, it seems that adapting the word vectors during the model training does not help. This is perhaps due to the limited training data for each tune, which may result in over-fitting. Another observation from the fourth row is that the global context supervision does improve the generation. Finally, the hybrid training model can deliver better performance than the model trained with data belonging to a single tune.

From these results, we obtain the best configuration for the attention model (global context with pre-trained word vectors).
In the reset of the paper, the attention model with this configuration is denoted by Attention-1, and the model trained by hybrid-tune training with $11$ tunes is denoted by Attention-11.

\subsection{Results in Phase 2}

In the second phase, we invited 24 experts\footnote{Again, these experts are all professional in Chinese Song Iambics. We invited more experts than in Phase 1 to compare different methods in a more accurate way.} to conduct a series of scoring evaluation. These experts were asked to rate the attention model (three different settings) and two comparative methods, using a 1-5 scale in terms of the three metrics considered in the first phase evaluation. The two comparative methods are the SMT-based model (SMT) used by~\cite{he2012generating} and the RNN Language Model (RNNLM) method proposed by~\cite{mikolov2010recurrent}; the three attention-based models are Attention-1 (tune-specific models trained with data of their own tunes), Attention-11 (tune-independent model trained by hybrid-tune training with $3,718$ Song iambics belonging to $11$ tunes), Attention-ALL (tune-independent model trained by hybrid-tune training with all the $15,001$ Song iambics). Finally the original Song iambics written by ancient poets are also involved in the rating.

Table~\ref{tab:bestresult1} presents the results. It can be seen that our model outperforms both the SMT-based and the RNNLM-based approaches that are often used in Chinese poem Generation task.
Note that in this experiment, the RNNLM-based approach is also based on LSTM, and so holds the same advantage as the attention model in learning long-distance patterns. However the RNN model tends to swift from the initial theme, hence causing inconsistency during the generation.

When comparing the different settings of the attention models, we see that Attention-11 outperforms the Attention-1 in general, which is consistent with the pair-wised evaluation in the first phase evaluation. This double confirms that hybrid-tune training helps. This conclusion, however, does not always hold for particular tunes. For example, for `Pusaman' the hybrid-tune training in fact causes performance reduction. This can be explained by the tradeoff between data sparsity and data consistency: although hybrid-tune training employs more data and tends to improve model training, the extra data might be inconsistent to the task, thus leading to performance loss. Fortunately, with more and more data, the benefit with the hybrid-tune training becomes predominant and the performance is significantly improved ultimately, as demonstrated by the rating scores with the Attention-ALL model.

Finally, we note that almost in all the evaluate tasks, the original human-written iambics beat those generated by machines. On one hand, this indicates that human are still superior in artistic activities, and on the other hand, it demonstrates from another angle that the participants of the evaluation are truly professional and can tell good and bad Song iambics. Interestingly, in the metric `Poeticness', the Attention-All model outperforms human. This is not surprising as computers can simply search vast candidate characters to ensure a rule-obeyed generation, however human artists put meaning and affection as the top priority, so sometimes break the rules.

To support the subjective evaluation, we show the BLEU results in Table~\ref{tab:bleu}. It can be seen that the BLEU results are highly consistent to the results of the subjective evaluation. A minor exception is that the SMT-based approach outperforms the RNNLM-based approach. A possible reason is that the BLEU metric focus on keyword co-occurrence rather than reasonable word sequence, which is therefore more amiable to the phrase-based SMT model that we used than RNN.

\subsection{Generation Example}

Finally we show an example Song iambics generated by the Attention-ALL model. The theme of this iambics is about sadness for the past time when the poet stands by the river.

\begin{table}[!htb]
\begin{center}
\begin{tabular}{|c|c|c|}
\hline
\begin{CJK*}{UTF8}{gbsn}
菩萨蛮
\end{CJK*}
\\
Pusaman\\
\begin{CJK*}{UTF8}{gbsn}
哀筝一弄湘江曲，
\end{CJK*}
\\
A sad melody flows with the wave,\\
\begin{CJK*}{UTF8}{gbsn}
风流水上人家绿。
\end{CJK*}
\\
But still everything seems to thrive.
\\
\begin{CJK*}{UTF8}{gbsn}
小艇子规啼，
\end{CJK*}
\\
A cuckoo sings over my little boat,\\
\begin{CJK*}{UTF8}{gbsn}
不堪春去时。
\end{CJK*}
\\
Hard to let spring go.\\
\begin{CJK*}{UTF8}{gbsn}
花前杨柳下，
\end{CJK*}
\\
Blossoms and green willows we found,\\
\begin{CJK*}{UTF8}{gbsn}
红叶满庭洒。
\end{CJK*}
\\
Soon will be maple leaves red on the ground.\\
\begin{CJK*}{UTF8}{gbsn}
月落尽成秋，
\end{CJK*}
\\
The moon is setting, the fall is coming,\\
\begin{CJK*}{UTF8}{gbsn}
愁思欲寄留。
\end{CJK*}
\\
But my longing for the past lingers around.\\
\hline
\end{tabular}
\end{center}
\caption{An example of Song iambics in tune 'Pusaman' generated by the attention model (Attention-ALL).}
\label{tab:Song}
\end{table}

\section{Conclusion}

This paper proposed an attention-based sequence-to-sequence learning approach for Chinese Song iambics generation.
Compared to several popular poetry generation methods, the new approach is simple in model structure, flexible in learning variable-length sentences and powerful in learning complex regulations. The subjective evaluation results show that with a large-scale hybrid training, the attention model can generate Song iambics pretty well. Although we can not beat human artists yet, the present result ($4.08$ V.S. $4.45$) is highly encouraging. With more data involved and continuous refinement of the model, it seems not impossible for machines to generate human-level iambics.

A future work will utilize other resources to improve Song iambics generation, e.g., Tang poetry. We will also apply this model to other forms of literary genres in Chinese, e.g., Han Fu, Yuan qu, and even novels.

\section*{Acknowledgments}

This research was supported by the National Science Foundation of China (NSFC) under the project No. 61371136,
and the MESTDC PhD Foundation Project No. 20130002120011. Many thanks to Bingjie Liu for the translation of the iambics in Table~\ref{tab:Songsample} and Table~\ref{tab:Song}. Also thanks to the experts who participated the evaluation, including Yinan Zhang, Zhou Tan, Siwei Li, Caizhi Chen, Zhihong Chen, Boyang Ni, Yufeng Hou, Yongsheng Li, Min Chen, Yunwei Du, Yaojun Luo, Haiying Li and many others.

\clearpage
\newpage
\bibliographystyle{named}
\bibliography{ijcai16}

\end{document}